# Utility - Probability Duality

Ali Abbas**,** < aliabbas@stanford.edu >
Lecturer,
*Department of Management Science and Engineering*
Stanford University

James E. Matheson, < jmatheson@smartorg.com >
Chairman, SmartOrg, Inc. and Consulting Professor,
*Department of Management Science and Engineering*
Stanford University

## Abstract

This paper introduces duality between probability distributions and utility functions. The primal problem is to maximize the expected utility over a set of probability distributions. To develop the dual problem, we scale the utility function between zero and one, so that it obeys the same mathematical properties as a (cumulative) probability function. We show that reversing the roles of the two functions in the expected utility formulation provides a natural "dual" problem. Many of the known results for the primal problem can be reinterpreted in the dual problem. For example, we introduce a new quantity, the aspiration equivalent, as the "dual" of the certain equivalent. The aspiration equivalent provides a new method for choosing between lotteries and a win-win situation for principal-agent delegation when used as a target. We also show several new dual results such as utility dominance relationships as dual to stochastic dominance relationships and introduce a new saddle-point method for allocating lotteries to decision makers.

# 1 – Introduction to duality

In the world of normative decision analysis, we model a decision situation with a utility function, over a variable $x$, representing the decision maker's preferences, and a set of probability distributions representing the decision maker's information about the lotteries he is facing. For any given lottery, the certain equivalent is defined as the $x$-amount for certain that the decision maker finds equivalent to that lottery. We can think of the certain equivalent as an equivalent step cumulative probability distribution, which provides the same expected utility as the original lottery. In this "primal" world, by maximizing the certain equivalent we maximize the expected utility of this (single) utility function over the set of lotteries.

We begin our development with the observation that if a utility function is normalized to range from zero to one, it behaves mathematically as a cumulative probability distribution (both are non-decreasing and range from zero to one). This leads us to think of a "dual" mirror-image world with the roles of probability distributions and utility functions interchanged. In the dual world, we calculate an $x$-equivalent by replacing the utility function with an equivalent step utility function, which has the same expected utility as the original utility function. Classically this step function is called an aspiration-level utility function, and the $x$-amount is called its aspiration level. We define this amount as the aspiration equivalent. Aspiration utility functions appear in many situations in practice. When managers are asked to meet performance targets, in effect they are asked to act with aspiration utility functions.

In a straightforward decision analysis, a decision maker chooses the best of a set of lotteries. Classically, the decision analyst asks the decision maker to choose the lottery with the highest certain equivalent. We refer to this problem as the "primal" problem. We show that the aspiration equivalent of the dual formulation provides a new method for choosing between lotteries, where the decision analyst would ask the decision maker to choose the lottery with the highest probability of meeting its aspiration. Decision analysts are used to the primal approach, however, if the aspiration equivalent is used as a target, the second approach may be more comfortable and persuasive for many real-word executives who routinely manage by setting targets and objectives for their subordinates.

We also show how the aspiration equivalent provides a unique win-win situation when used as a target in principal-agent delegation. In this setting, the agent maximizes the probability of exceeding his target when choosing between lotteries, and at the same time, maximizes the principal's expected utility.

The dual formulation provides solutions to many problems that are not usually treated by classical decision analysis or expected utility theory. For example, we consider situations where a manager suddenly faces worse (or better) lotteries than anticipated. Should he continue to hold the previous targets? We illustrate how our formulation provides a method to update the targets and is consistent with the decision maker's risk preference. We also show that consistency among risk attitudes implies that higher aspiration targets go naturally with higher risk tolerances.

The dual formulation we discuss leads naturally to the introduction of the "dual" decision problem, which selects the decision maker with the highest aspiration for a given lottery. We show how the solution to this formulation can be simplified by observing certain dominance relations between utility functions, analogous to stochastic dominance relations between cumulative probability distributions. We also discuss the problem of allocating several lotteries to several decision makers, which becomes a game theoretic formulation. We introduce a new saddle-point method of allocating lotteries to decision makers. Finally we show how many well known theorems and properties from the primal view are translated into new analogous ones from the dual view.

The utility-probability duality that we develop in this paper has not been seen in our search of the literature. Berhold (1973) rescales probability distributions to obtain convenient expressions for utility functions, but he does not introduce duality. Castagnoli and LiCalzi (1996) interpret a normalized utility function as a probability distribution of a hypothetical lottery that is independent of the lotteries being faced by the decision maker. The expected utility may then be interpreted as the probability that the outcome of this hypothetical lottery is less than the outcome of the decision maker's lottery. Bordley and LiCalzi (2000) also discuss aspiration levels and propose that one should select an action that maximizes the probability of meeting an uncertain target, which is equivalent to maximizing expected utility. In contrast we interpret the normalized utility function as simply representing preference probabilities derived from the typical indifference preference assessments used to obtain a utility function. Thus in our work, we maintain the separation of beliefs about the likelihood of events from preferences over the potential results of those events.

## 2 – The Aspiration Equivalent

Assuming a continuous and increasing utility function, the certain equivalent is defined by replacing a given cumulative probability function with a step cumulative distribution function yielding the same expected utility. The certain equivalent is the point at which this step occurs. The value of the certain equivalent depends on both the utility function and probability distribution that have been specified.

In the mirror world of duality, assuming a continuous and increasing cumulative probability function, we define an aspiration equivalent by replacing the utility function with a step utility function yielding the same expected utility. Similarly, the aspiration equivalent is the point at which this step occurs, and its value also depends on both the utility function and probability distribution that have been specified. This step utility function is shown in figure 1, and is known as an aspiration utility function.

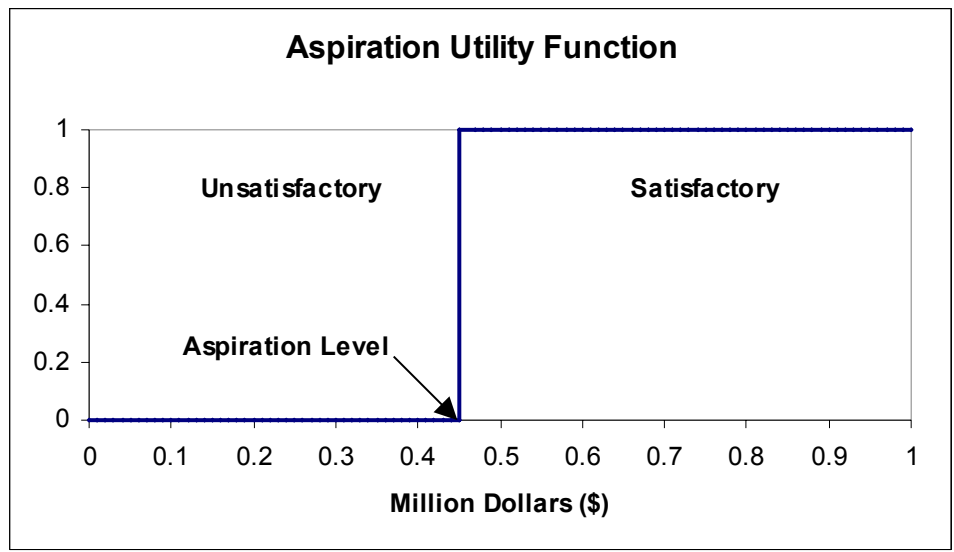

Figure 1 – Aspiration utility function

The aspiration utility function has had a large share of literature coverage. The point at which its step occurs is traditionally known as the aspiration level. Simon (1955) defined the aspiration level as a "satisfactory alternative" and suggested that individuals could simplify decision problems by having binary goals: satisfactory if the outcome is above the aspiration level or unsatisfactory if it is below it.

> However, there are certain dynamic considerations, having a good psychological foundation that we should introduce at this point. Let us consider, instead of a single static choice situation, a sequence of such situations. The *aspiration level*, which defines a satisfactory alternative, may change from point to point in this sequence of trials. [Italics Simon's]

Simon suggested that the aspiration level could change with changing "choice situations" or with different lotteries faced by the individual. In this paper, we develop a mathematical formulation to define an aspiration level, which we call the aspiration equivalent, and show how to adapt it based on changes in the lottery that the individual is facing.

## Foundations of Utility-Probability Duality

We remind the reader that if a decision maker follows the axioms of decision analysis, then any complex lottery can be reduced to an equivalent lottery with only two outcomes, the best and worst. This is achieved by introducing a utility function characterized by preference probabilities. Furthermore, the probability of getting the best outcome in this equivalent two-outcome lottery is equal to the expected utility of the lottery, which can be defined as

$$U(\tilde{x}) \triangleq Expected\ Utility = \int_{-\infty}^{\infty} U(x)dF(x) = \int_{-\infty}^{\infty} f(x)U(x)dx, \qquad (1)$$

where $F(x)$ is the cumulative probability function of the lottery, $f(x)$ is the probability density function, $U(x)$ is the decision maker's normalized utility function, and $\tilde{x}$ is the certain equivalent of the lottery, which is unique if $U(x)$ is a continuous increasing function at $\tilde{x}$.

Using the rule of integration by parts we can express the expected utility as

$$Expected\ Utility = \int_{-\infty}^{\infty} U(x)dF(x)$$

$$= U(x)F(x)\big|_{-\infty}^{\infty} - \int_{-\infty}^{\infty} F(x)dU(x) \tag{2}$$

$$= 1 - \int_{-\infty}^{\infty} F(x)dU(x)$$

$$\triangleq 1 - Expected\ Disutility$$

We define the last term on the right hand side as the expected disutility. Now we define the aspiration equivalent, $\hat{x}$, by the equation

$$F(\hat{x}) \triangleq Expected\ Disutility = \int_{-\infty}^{\infty} F(x)dU(x) = \int_{-\infty}^{\infty} u(x)F(x)dx. \tag{3}$$

where $F(x)$ is the cumulative probability distribution of the lottery and $u(x)$ is the decision maker's utility density function. A utility density function, $u(x)$, is defined as the derivative of a normalized utility function, $U(x)$ (Abbas, 2002). A utility density function is non-negative (due to the non-decreasing property of utility functions) and integrates to unity. The aspiration equivalent is unique if $F(x)$ is a continuous increasing function at $\hat{x}$.

Comparing the expression for the expected utility of equation (1) and expected disutility in equation (3), we note that the probability and utility functions and their densities have been interchanged in the integral.

Combining equations (1), (2) and (3), for any normalized utility function we have

$$Expected\ Utility + Expected\ Disutility\ =\ 1$$

$$U(\tilde{x}) + F(\hat{x}) = 1 \tag{4}$$

Equation (4) presents the fundamental identity of utility-probability duality and provides several new interpretations. We note that, since the expected utility is the probability of getting the best prospect in the equivalent two-outcome lottery, and the expected disutility is the probability of getting the worst prospect in that same equivalent lottery. The problem of choosing the lottery that has the highest expected utility is thus equivalent to the problem of choosing the lottery that has the lowest expected disutility.

Furthermore, given an uncertain lottery and a utility function, the certain equivalent of the lottery is the value of an equivalent deterministic lottery that has a step cumulative distribution at the value of the certain equivalent. Analogously, the aspiration equivalent of a utility function is the value at which an equivalent step utility function jumps from zero to one. If you replace the utility function with this aspiration level utility function, while leaving the probability function unchanged, you will get the same expected utility. Now let us re-arrange equation (4) as follows

$$Expected\ Utility =\ U(\tilde{x}) = 1 - F(\hat{x}) = G(\hat{x}), \qquad (5)$$

where $G(x) \triangleq 1 - F(x)$ is the excess distribution function.

From equation (5) we have a new interpretation for the expected utility of a lottery in terms of the aspiration equivalent. The expected utility of a lottery is the probability that the outcome of the lottery exceeds its aspiration equivalent that is determined by equation (3). This provides us with a new method for choosing between lotteries: *we choose the lottery that has the highest probability of meeting its aspiration equivalent*.

Consequently the aspiration equivalent is the point that divides the x-axis of the cumulative probability distribution into two portions, the probability of the portion to the right is numerically equal to the expected utility of the lottery, or the probability of the most desirable outcome, b, and the probability of the portion to the left is the expected disutility or probability for the least desirable outcome, a. This is shown in figure 2 below.

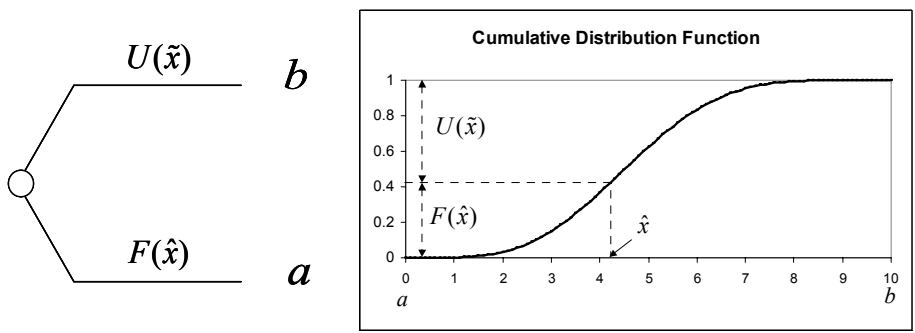

Figure 2 – The aspiration equivalent splits the cumulative distribution function into the probabilities of the best and worst prospects of the equivalent two-outcome lottery.

  Utility Probability Duality 10-27-03.doc

The following example presents a direct numerical calculation of the aspiration equivalent and the certain equivalent for a given lottery and a given utility function.

**Example – Aspiration Equivalent vs. Certain Equivalent**

A decision maker with an exponential utility function and a risk aversion coefficient, $\gamma = 0.03$, over the domain [$0, $200], is facing a symmetric triangular probability density function over the same domain as shown in figure 3 below:

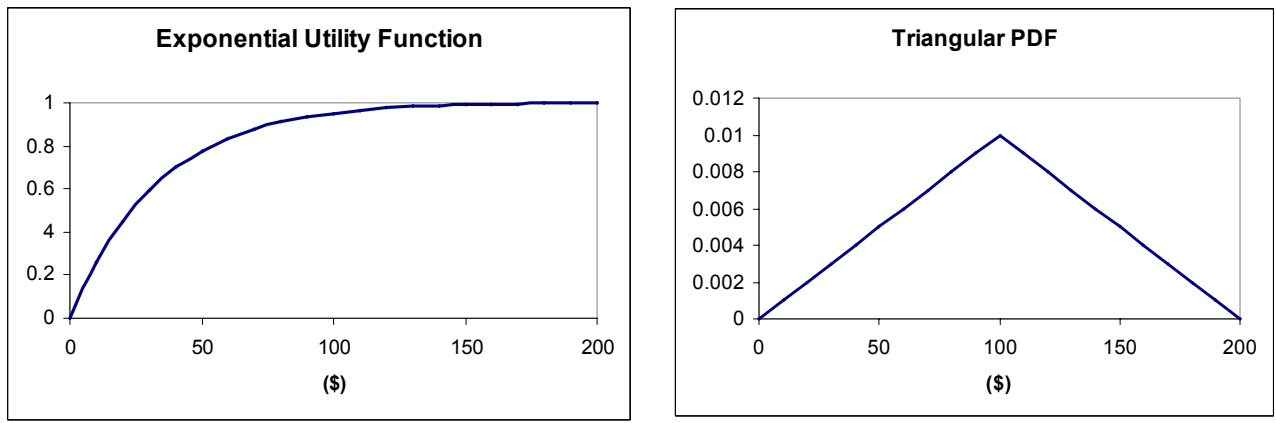

Figure 3 – Exponential utility and triangular probability example

The equation for the normalized utility function is given as:

$$U(x) = \frac{1 - e^{-\gamma x}}{1 - e^{-200\gamma}}, 0 \le x \le 200 \tag{6}$$

where, $\gamma$ is the risk aversion coefficient.

By direct integration, we can calculate the certain equivalent and aspiration equivalent of the decision maker for the given lottery and given risk aversion coefficient.

$$ExpectedUtility = \int_0^{100} (\frac{x}{100}) \frac{1 - e^{-\gamma x}}{1 - e^{-200\gamma}} dx + \int_{100}^{200} \frac{(200 - x)}{100} \frac{1 - e^{-\gamma x}}{1 - e^{-200\gamma}} dx = 0.899 \tag{7}$$

On figure 4, we have superimposed the excess distribution function, $G(x)$, and the utility function, $U(x)$. If we draw a horizontal line at the value of the expected utility, it intersects $G(x)$ at the aspiration equivalent and $U(x)$ at the certain equivalent. Traditional decision analysts consider a single utility function and many lotteries so it is natural to project expected utility through the single utility function obtaining a certain equivalent for each lottery. However, for a dual problem with a single lottery and many utility functions, it is natural to project expected utility through the single lottery obtaining an aspiration equivalent for each utility function. We will discuss this dual problem further in section (5).

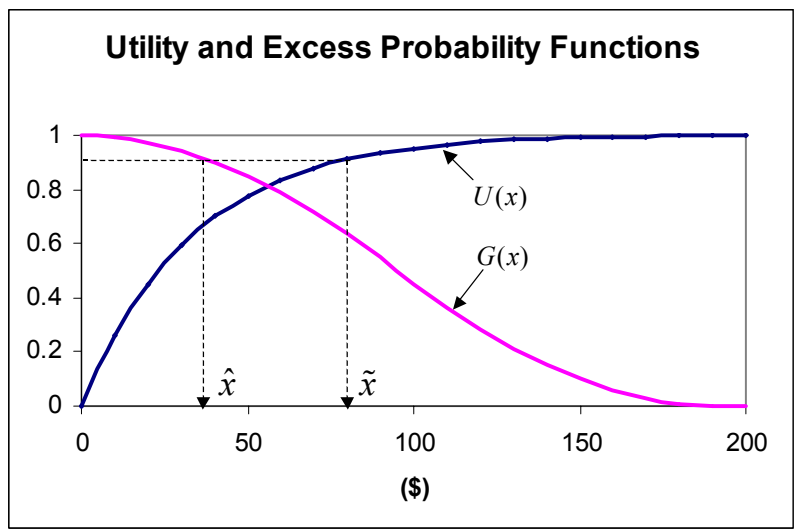

Figure 4. Certain equivalent and aspiration equivalent calculation.

We point out here that aspiration levels have been treated by many authors. For example, Payne, Laughum, and Crum, (1980) show experimental evidence for the effects of aspiration levels on risky choice behavior, Lant (1992) performed experimental studies to show empirical evidence of aspiration adaptation with performance feedback, and March (1988) discussed variation of risk preferences with aspiration adaptation. In section (3) we show that the aspiration equivalent is an appropriate target to set in delegation from a principal to an agent and explore aspiration adaptation in the face of new lotteries. First, we discuss the following numerical example to gain some insights into the behavior of the aspiration equivalent with various degrees of risk preference and relate our results to the behavioral literature.

**Example – Sensitivity to Risk Aversion Coefficient**

Now we discuss the variation of both certain equivalent and aspiration equivalent with the risk attitude of the decision maker. Figure 5 shows a sensitivity analysis of both certain equivalent and aspiration equivalent to the risk aversion coefficient, $\gamma$ and the triangular lottery of the previous example.

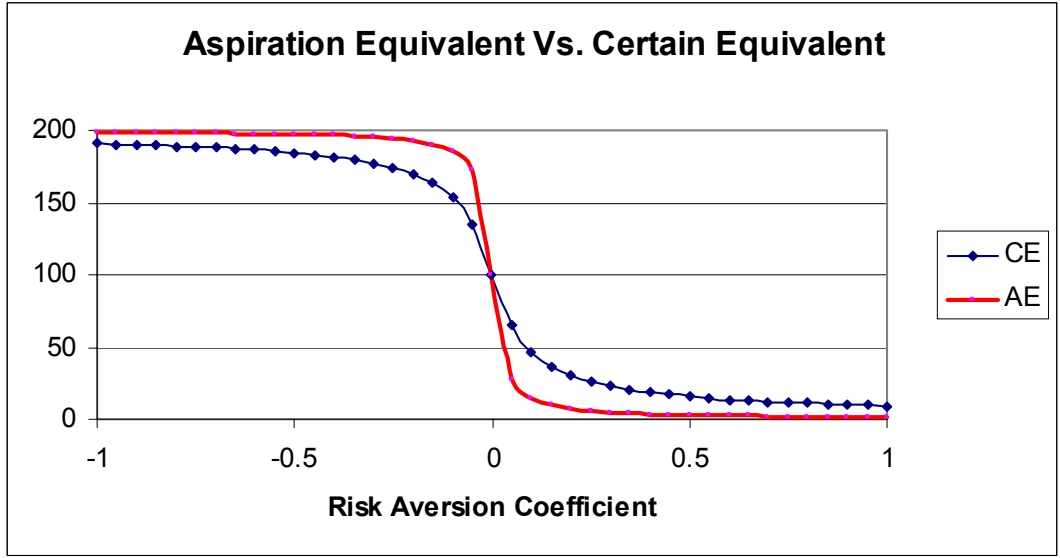

Figure 5 – Sensitivity analysis of the certain equivalent and the aspiration equivalent to the risk aversion coefficient.

The first observation we note from figure 5 is that both the aspiration equivalent and the certain equivalent increase for a given lottery as the risk aversion coefficient, $\gamma$, decreases, and decrease as $\gamma$ increases. This relation illustrates that high aspiration equivalents are incompatible with low risk tolerances. The relation between the aspiration equivalent and the risk aversion coefficient can be derived mathematically for any general lottery as follows. For a decision maker that has an exponential utility function on the domain [a, b], the utility density function is

$$u(x) = \frac{\gamma e^{-\gamma x}}{e^{-\gamma a} - e^{-\gamma b}}, \; a \le x \le b \tag{8}$$

As $\gamma \to +\infty$, $u(x) = \dfrac{\gamma e^{-\gamma x}}{e^{-\gamma a} - e^{-\gamma b}} \to \delta(x-a)$, $\tag{9}$

where $\delta(x-a)$ is an impulse delta function at $x = a$, which corresponds to a step (aspiration) utility function at that value of $x$.

$$F(\hat{x}) = \lim_{\gamma \to \infty} \int_a^b \frac{\gamma}{e^{-\gamma(a-x)} - e^{-\gamma(b-x)}} F(x)dx = F(a) = 0 \quad \Rightarrow \hat{x} \to \text{lower bound of lottery} \qquad (10)$$

As $\gamma \to -\infty$, $u(x) = \frac{\gamma e^{-\gamma x}}{e^{-\gamma a} - e^{-\gamma b}} \to \delta(x-b)$ \qquad (11)

$$F(\hat{x}) = \lim_{\gamma \to -\infty} \int_a^b \frac{\gamma}{e^{-\gamma(a-x)} - e^{-\gamma(b-x)}} F(x)dx = F(b) = 1 \quad \Rightarrow \hat{x} \to \text{upper bound of lottery} \qquad (12)$$

If the decision maker is facing a lottery and her aspiration equivalent is set to the lower bound, she is effectively acting with risk-averse behavior. On the other hand, if a decision maker is facing a lottery and her aspiration equivalent is set close to the upper bound of the lottery, she is effectively acting with risk-seeking behavior.

We can see also from figure 5 that when the risk aversion coefficient, $\gamma$, approaches zero, both certain equivalent and aspiration equivalent converge to the expected value of the lottery (this is a general result for any symmetric lottery). As the decision maker becomes more risk averse the aspiration equivalent falls below the certain equivalent, and, as the risk aversion coefficient increases, both terms approach the lower bound of the lottery. This also means that the risk-averse decision maker is more likely to exceed his aspiration equivalent than his certain equivalent – the aspiration equivalent is an easier target. On the other hand, as the decision maker becomes risk seeking, the value of the aspiration equivalent exceeds the certain equivalent and, as he becomes more risk seeking, both terms approach the upper bound of the lottery. This means that a risk seeking decision maker is less likely to exceed his aspiration equivalent than his certain equivalent – the aspiration equivalent is a more difficult target. This result is in harmony with the discussion of shifts of reference points in Kahneman and Tversky's prospect theory (1979):

> There are situations in which gains and losses are coded relative to an expectation or *aspiration level* that differs from the status quo. For example an unexpected tax withdrawal from a monthly paycheck is experienced as a loss, not as a reduced gain. Similarly, an entrepreneur who is weathering a slump with greater success than his competitors may interpret a small loss as a gain, relative to the larger loss he had reason to expect…a person who has not made peace with his losses is likely to accept gambles that would be unacceptable otherwise. The well known observation that the tendency to bet on long shots increases in the course of the betting day *provides some support for the hypothesis that a failure to adapt to losses or to attain an expected gain induces risk seeking.* [Italics Kahneman and Tversky's]

Kahneman and Tversky show that high aspirations, that are difficult to realize, induce risk-seeking behavior. In our formulation, setting high aspirations, that are difficult to realize, is associated with risk seeking behavior.

From the results of figure 5 we also see that *for any given lottery and given aspiration equivalent, there exists an exponential utility function that provides the same expected utility (and aspiration equivalent) and thus determines a unique "effective risk-aversion coefficient".* This is an important result that relates targets and aspirations to risk aversion coefficients. March and Shapira (1992) discuss related work in behavioral settings that relates aspirations to risk preference. Using equations (3) and (8) we have

$$F(\hat{x}) = \int_a^b u(x)F(x)dx = \int_a^b \frac{\gamma e^{-\gamma x}}{e^{-\gamma a} - e^{-\gamma b}} F(x)dx. \tag{13}$$

Using equation (13) we can now solve for the effective risk aversion coefficient, $\gamma_{eff} = \gamma(F, \hat{x})$, for any given lottery, $F$, and aspiration equivalent, $\hat{x}$.

## 3–Principal-Agent Delegation

In organizational settings, such as corporate management, executives sometimes ask managers to meet fixed targets; they are, in effect, asking them to act with a fixed aspiration level. A behavioral rationale for setting targets is a translation from the executive world of strategic decision making to an operational world of managing with targets, using tools such as management by objectives (MBO) and balanced scorecards. We can think of this situation as a principal-agent delegation problem, where the agent is rewarded when he meets his target. This incentive scheme thus leads the agent to give priority to the actions that maximize the probability of meeting this target. This behavior often leads the agent to choices that do not maximize the principal's expected utility. We will explore how a target can be set to motivate the agent to maximize the principal's expected utility while, at the same time, maximize the probability of meeting his target. We will assume in this section that both the principal and the agent agree on probability distributions, or equivalently, that the principal trusts the agent's probability assessments.

We will require three basic desiderata of this target: (i) that it can be used to choose between lotteries; (ii) that when used for delegation from a principal to an agent, who maximizes his probability of exceeding his target, it maximizes the principal's expected utility, and (iii) that there exists some normative and transparent method, consistent with the principal's preferences (utility function), to update the target if the lottery changes.

Let us now consider three possible choices of such targets: fractiles of a distribution, the certain equivalent, and the aspiration equivalent. The first desideratum implicitly requires the target to depend on the principal's utility function; otherwise it cannot be used to choose between lotteries. Fractiles of a lottery fail this basic desideratum since they do not include information about the decision-maker's utility function.

The certain equivalent, on the other hand, incorporates information about the lottery and the decision-maker's utility function, and expected utility theory suggests that a decision maker should choose the lottery that maximizes his certain equivalent. The certain equivalent thus passes the first desideratum of choosing between lotteries. Now we consider its use in delegation from the principal to the agent. In a principal-agent setting, choosing the lottery with the highest probability of meeting its certain equivalent (if used as a target) is not equivalent to the lottery that maximizes the principal's expected utility. The certain equivalent target thus fails the second desideratum.

Now we consider the aspiration equivalent as a target. The aspiration equivalent incorporates information about both the lottery and the decision maker's utility function. From equation (5), we see that it can be used to choose between lotteries if we select the lottery that has the highest probability of meeting its aspiration equivalent. Now we discuss the role of the aspiration equivalent in delegation. If the principal delegates the aspiration equivalent to the agent, and the agent chooses the lottery that has the highest probability of meeting its target, then he will choose the lottery that has the highest expected utility for the principal. This choice creates a win-win situation for both the principal and the agent. Furthermore, if the agent discovers a new lottery that is better for the principal, the agent can calculate a new target that he has a higher probability of exceeding. The agent will therefore choose the new lottery in the best interest of the principal.

The aspiration equivalent thus satisfies the first two desiderata (delegation and choice among lotteries). In the next section we will show that the aspiration equivalent provides a convenient method to update the target, which satisfies the third desideratum above.

## 4– Adjusting Aspirations to a New Forecast

This section shows by example how a constant effective risk aversion provides a natural method to adjust aspirations. Consider the following situation.

A project manager is asked to meet a target of $3 Million revenue for a project, which has a distribution over the range 0 to $ 10 million given by a scaled Beta (4,6). Assume that the initial target has been set using an aspiration equivalent of $3M, which gives the manager an 85% chance of meeting his goal. The agent uncovers a new alternative that increases the revenue forecast to a scaled Beta (6,3). Now the manager has a "windfall" of nearly 100% chance of meeting his goal. These two distributions are shown in figure 6 below. What should the new target be?

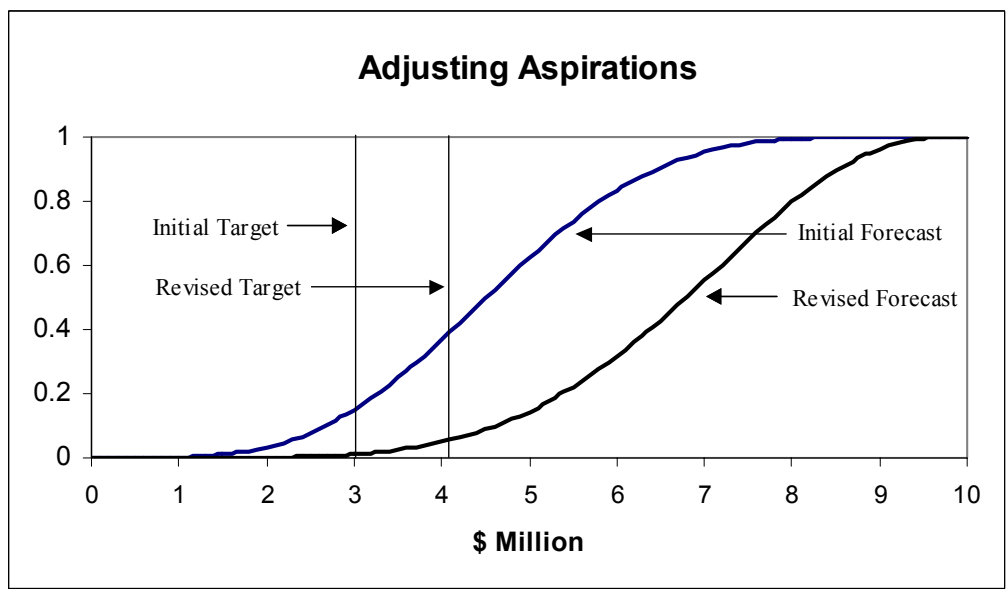

Figure 6 – Adjusting aspirations in the face of a new forecast.

In a business delegation context, it is not reasonable to have the same target for the new distribution. The target would be too easily achieved given the second forecast, and operational managers might not try hard enough to achieve the high end of the forecast. However, if organization desires the same effective risk tolerance in both cases, we can calculate the revised target as follows.

First we calculate the effective risk tolerance that would produce this target. Solving equation (13) for the unique risk aversion coefficient, where $\hat{x} = \$3M$ and $F(\hat{x}) = 0.15$, we obtain $\gamma_{eff} = 1/\$2.3M$. In a corporate setting the agent would interpret this value as the corporate risk aversion coefficient. Consistency in risk attitude requires using this value to calculate the aspiration equivalent of the new lottery. Using the revised forecast, the new aspiration equivalent can be calculated using equation (3) as $4.1M. The manager now has increased his probability of meeting the target from 85% to 95%. Furthermore, the principal's expected utility has increased from 0.85 to 0.95, illustrating the win-win situation.

## 5–Selecting Decision Makers

In classical decision analysis, we are faced with several lotteries and are interested in the one that has the highest certain equivalent or highest expected utility. We have also seen that this problem is equivalent to choosing the lottery that has the highest probability of meeting its aspiration equivalent.

Now we discuss the dual problem, which is selecting the decision maker (or utility function) over all decision makers (utility functions) that will provide the highest aspiration equivalent for a given lottery. For example, an entrepreneur may be interested in selecting a CEO who will have the highest aspirations (or target) for the results of his business. This CEO will also have the lowest probability of meeting his target and receiving his bonus.

By reversing the role of the utility and probability functions and introducing $n$ utility functions $U_j$, we now formulate the dual problem:

$$\begin{array}{ccc} & \text{PrimalProblem} & \text{Dual Problem} \\ Utility: & \max_i \int U(x)dF_i(x) & \min_j \int U_j(x)dF(x), \\ & \multicolumn{2}{c}{\text{or equivalently}} \\ Disutility: & \min_i \int F_i(x)dU(x) & \max_j \int F(x)dU_j(x) \end{array} \tag{14}$$

The dual problem is thus either minimizing expected utility or maximizing the expected disutility over all possible utility functions (decision makers involved). The upper left expression maximizes expected utility in the primal world while the lower right expression maximizes expected utility in the mirror dual world. By maximizing expected utility over lotteries in the primal problem, we also maximize the certain equivalent, *for a single fixed utility function*. By minimizing expected utility over utility functions in the dual problem, we also maximize the aspiration equivalent, *for a single fixed probability function*.

**Example: Selecting Fund Managers**

A brokerage firm would like to report the highest target for its new fund whose projections are represented by a scaled Beta (2,8) distribution on a scale of 0 to $ 10 million. The firm is interested in selecting one of three brand fund managers to manage this fund. The brand fund managers have risk tolerances of 3.3, 1.6, and 1.1 million dollars, corresponding to risk aversion coefficients of 3, 6, and 9 (x $10^{-7}$) respectively. Table 1 shows the expected utility, disutility, certain equivalent, and aspiration equivalent calculations for each fund manager. Note the sum of the corresponding cells in tables (a) and (b) must be unity.

| EU | Risk Aversion Gamma | | |
|---|---|---|---|
| | 3 | 6 | 9 |
| Beta(2,8) | 0.45 | 0.64 | 0.75 |

(a)

| EDU | Risk Aversion Gamma | | |
|---|---|---|---|
| | 3 | 6 | 9 |
| Beta(2,8) | 0.55 | 0.36 | 0.25 |

(b)

| CE( $ 10 M) | Risk Aversion Gamma | | |
|---|---|---|---|
| | 3 | 6 | 9 |
| Beta(2,8) | 0.19 | 0.17 | 0.16 |

(c)

| AE( $ 10 M) | Risk Aversion Gamma | | |
|---|---|---|---|
| | 3 | 6 | 9 |
| Beta(2,8) | 0.20 | 0.14 | 0.11 |

(d)

Table 1 – Numerical examples for three fund managers

The fund manager who has the lowest risk aversion coefficient (the highest risk tolerance) has the highest aspiration equivalent and therefore the lowest probability of reaching his target and receiving his bonus. He will set a target for the fund equal to $2 million.

## 6–Simultaneously selecting decision makers and lotteries

### The Saddle-Point Property

Having presented the dual problem of choosing between utility functions for a given lottery, we now consider the problem of having a set of probability functions $F_i$ and a set of utility functions $U_j$. The expected utility (or disutility) may be regarded as a scalar bilinear form of vectors from the two sets of vectors, $F_i$ and $U_j$. In 1944, von Neumann and Morgenstern recognized the possible existence of a saddle point for such problems. Their book explores the conditions under which the saddle-point property exists for both pure strategies and mixed strategies. To relate our work to theirs, one can think of two players, Dr. Jekyll and Mr. Hyde, one choosing from the set of probability functions and the other choosing from the set of utility functions. Dr. Jekyll is attempting to maximize his expected utility while his evil mirror image is trying to minimize it. If we use our interpretation of the expected utility as the probability of meeting the aspiration equivalent, then Dr. Jekyll is choosing the lottery that has the highest chance of meeting its aspiration, while Dr. Hyde is choosing the utility function that will provide the lowest chance of meeting the aspiration equivalent for the lottery that Dr. Jekyll has chosen. Thus Dr. Jekyll and Mr. Hyde are in a zero-sum game. Each selects the lottery (or utility function) that minimizes the maximum damage the other can do. At the saddle point, neither party can improve his situation by making another selection. If a saddle point exists, the duality equations below hold with equality.

$$
\begin{array}{cc}
\text{Primal Problem} & \text{Dual Problem} \\
\end{array}
$$

$$Utility: \quad \min_j \max_i \int U_j(x)dF_i(x) \geq \max_i \min_j \int U_j(x)dF_i(x),$$

$$\text{or equivalently}$$

$$Disutility: \quad \max_j \min_i \int F_i(x)dU_j(x) \leq \min_i \max_j \int F_i(x)dU_j(x)$$

(15)

Let us now demonstrate the saddle point idea through a numerical example, and discuss a saddle point method of assigning lotteries to utility functions.

**Example: Assigning Fund Managers and Funds**

Consider the case of the brokerage firm that we discussed earlier. This time the firm has a set of three funds and is interested in assigning one to each of the three name-brand fund managers. The firm would like to publicize high aspiration equivalents while minimizing the probability of paying bonuses to the agents. We will use three probability density functions for the funds, two of which have the same mean, Beta (2,8) and Beta (3,5) as well as another probability density function, Beta (4,8), which has stochastic dominance with respect to the other two. We will assume the managers have the same three exponential utility functions with risk aversion coefficients, $\gamma = 3, 6,$ and $9 \, (x \, 10^{-7})$. The results are shown in table (2) below.

| EU | Gamma | | |
|---|---|---|---|
| | 3 | 6 | 9 |
| Beta(2,8) | 0.45 | 0.64 | 0.75 |
| Beta(3,12) | 0.46 | 0.66 | 0.78 |
| Beta(4,8) | 0.64 | 0.83 | 0.91 |

| EDU | Gamma | | |
|---|---|---|---|
| | 3 | 6 | 9 |
| Beta(2,8) | 0.55 | 0.36 | 0.25 |
| Beta(3,12) | 0.54 | 0.34 | 0.22 |
| Beta(4,8) | 0.36 | 0.17 | 0.09 |

(a) Expected Utility  (b) Expected Disutility

| CE | Gamma | | |
|---|---|---|---|
| | 3.00 | 6.00 | 9.00 |
| Beta(2,8) | 0.19 | 0.17 | 0.16 |
| Beta(3,12) | 0.19 | 0.18 | 0.17 |
| Beta(4,8) | 0.31 | 0.29 | 0.27 |

| AE | Gamma | | |
|---|---|---|---|
| | 3.00 | 6.00 | 9.00 |
| Beta(2,8) | 0.20 | 0.14 | 0.11 |
| Beta(3,12) | 0.20 | 0.15 | 0.12 |
| Beta(4,8) | 0.28 | 0.21 | 0.16 |

(c) Certain Equivalent  (d) Aspiration Equivalent

Table 2 – Numerical examples for duality in the continuous case

In each column we can maximize the expected utility over all possible probability distributions. For example, the maximum cell in the first column of table 2 (a), which is in the third row, is 0.64. This is also the minimum of all of the column maximums. Now we take the minimum across the cells in the rows, and we find that the maximum of these minima is in the third row, namely 0.64. These two results are equal, illustrating the saddle-point property. Table 2(b) shows the corresponding expected disutility. The saddle-point property also holds for tables (b), (c), and (d).

This suggests a "saddle-point allocation" method of assigning lotteries to utility functions. The firm could calculate the initial saddle point, pair that lottery and recipient; then remove that pair from the two sets and repeat the calculation for the N-1 dimensional problem. Repeating this process will assign all pairs. At each stage the assignment maximizes the aspiration equivalent across recipients and the certain equivalent across lotteries.

If we apply the saddle-point allocation method to this problem, we would first match lottery 3 with utility 3, then match lottery 2 with utility 2, and finally match lottery 1 with utility 1. If the company uses the saddle-point allocation method, for each selection the brokerage firm could not do better in terms of the aspiration level, which they will advertise, by selecting another fund manager, and the selected fund manager could not achieve a higher certain equivalent by managing another sector. In this particular case, the saddle-point allocation also maximizes the sum of the certain equivalents and the sum of the aspiration equivalents over any other allocation.

Another application of the saddle point allocation method is in a general donor problem. Suppose a donor has an N-dimensional allocation problem of rewarding N lotteries to N recipients, one lottery per recipient. The utilities could either represent the beneficiaries' preferences or represent the donor's own preferences for the beneficiaries' outcomes. The donor could use the saddle-point allocation method to allocate the lotteries to the recipients.

## 7–The General Isomorphism

Many of the usual results for certain equivalents can also be translated (by the duality isomorphism) into similar results for the aspiration equivalent (since it is just a certain equivalent in the dual world). Because of space limitation we shall mention only a few results here.

### Power-Series Expansions

The certain equivalent of a lottery can be approximated by the first two terms of a Taylor expansion as follows

$$Certain\ Equivalent \cong First\ moment\ of\ lottery - \frac{1}{2}\frac{Second\ Central\ moment\ of\ lottery}{Risk\ Tolerance}, \quad (16)$$

where the $Risk\ Tolerance = -\dfrac{u(x)}{u^{'}(x)} = -\dfrac{U^{'}(x)}{U^{''}(x)}$ , which is tolerance to the spread of the probability function about its mean. It is well known that equation (16) is exact for decision makers with exponential utility functions facing Gaussian probability distributions.

Risk is a measure of spread of a probability function as measured by its second central moment. In an analogous manner, we define the spread of a utility function around its second central moment as simply the "spread". We can now write an approximate expression for the aspiration equivalent as

$$Aspiration\ Equivalent \cong First\ Moment\ of\ utility - \frac{1}{2}\frac{Second\ Central\ Moment\ of\ utility}{Spread\ Tolerance}, \quad (17)$$

where we define the $Spread\ Tolerance \triangleq -\dfrac{f(x)}{f^{'}(x)} = -\dfrac{F^{'}(x)}{F^{''}(x)}$ , which is tolerance to the spread about the first moment of the utility function.

By analogy with equation (16), equation (17) is exact for a decision maker with a Gaussian utility function and facing an exponential lottery. Equation (17) shows that as a first approximation the aspiration equivalent is determined by the first moment of the utility function. If we include other terms in the expansion, then the second central moment of the utility function and the spread tolerance also come into play.

Note from equation (17) that the first moment of a utility function is thus a first order approximation for the aspiration equivalent. If the lottery is uniform, $F^{''}(x) = 0$ and the spread tolerance is infinite. The aspiration equivalent is then exactly equal to the first moment of the utility function. We can thus think of the first moment of a utility function as a measure of aspiration that gets updated based on the lottery that the decision maker is facing. In (Abbas, 2002), we show that knowledge of the first moment of the utility function over a positive domain leads to an exponential maximum entropy utility function where the first moment is equal to the risk tolerance.

The difference between the first moment of the lottery and the certain equivalent is generally known as the risk premium. By analogy we can define the difference between the aspiration equivalent and the first moment of the utility function as the aspiration premium.

$$Aspiration\ Equivalent = first\ moment\ of\ utility - Spread\ premium \qquad (18)$$

The larger the value of the variance of the utility density function, the more significant is the effect of the spread tolerance on the aspiration premium. Furthermore, if the cumulative distribution of the lottery is strictly concave (the probability density function is monotonically decreasing), the spread premium is negative. If the cumulative distribution is convex, on the other hand, (the probability density function is monotonically increasing) and the spread premium is positive.

If both probability and utility functions are symmetric about the same point, at this point the spread tolerance is zero. The aspiration equivalent is therefore equal to the first moment of the utility function and lies at its inflection point. The decision maker is thus risk seeking below the aspiration equivalent and risk averse above it. This result is consistent with Kahneman and Tversky's prospect theory and is also consistent with Heath, Larrick, and Wu (1999) who suggest that the inflection point in an S-shaped utility function can be interpreted as a target or goal.

As a special case, and by direct duality with certain equivalents, we can draw on the results of Howard (1971) to derive an expression for the aspiration equivalent of an exponential lottery in terms of the cumulants of the utility function as

$$Aspiration\ Equivalent = \sum_{k=1}^{\infty} \frac{{}^{k}u}{k!}(-\lambda)^{k} \qquad (19)$$

where ${}^{k}u$ is the $k^{th}$ cumulant of the utility density function.

Similarly many Laplace transform formulas for certain equivalents with exponential utility functions go over into Laplace transform formulas for aspiration equivalents with exponential probability distributions.

Now we will illustrate the use of the approximation formula (17) for the aspiration equivalent.

**Example – Aspiration Equivalent Approximation**

A decision maker with an exponential utility function has a risk tolerance of $\rho = \dfrac{1}{\gamma} = $M 0.2. He is facing a lottery whose probability density function is a scaled Beta (2,3) from \$0 to \$1 million. The exact calculation of the aspiration equivalent is \$M 0.211. Let us now use the approximate formula.

The first moment of an exponential utility density function over the given domain is

$$\int_0^1 x u(x) dx = \int_0^1 x \frac{\gamma e^{-\gamma x}}{1 - e^{-\gamma}} dx = \$\text{M } 0.19 \tag{20}$$

We see again that the first moment of a utility function can be interpreted as the aspiration equivalent of a uniform lottery. If in this example, the exponential utility function is defined on the positive domain, the mean is simply the risk tolerance and in this case is \$M 0.2.

The second moment is given as

$$\int_0^1 x^2 u(x) dx = \int_0^1 x^2 \frac{\gamma e^{-\gamma x}}{1 - e^{-\gamma}} dx = 0.069 \tag{21}$$

The variance in the utility density is thus equal to 0.033.

The probability density function of a Beta (2,3) has the form $Beta(2,3) = 12\, x(1-x)^2$. The spread tolerance, calculated at the first moment of the utility density function, is -\$2.5. This leads to an approximation for the aspiration equivalent equal to \$M 0.207.

**Utility Dominance**

Stochastic dominance conditions simplify the analysis of the primal problem of maximizing expected utility over probability distributions. If the conditions are met, they imply that the dominating distribution will have the higher expected utility (and certain equivalent) for a matching class of utility functions. Many classes of stochastic dominance are presented in Levy (1998), and we will generally parallel his style here. By the duality isomorphism, utility dominance conditions simplify the analysis of the dual problem of maximizing expected disutility over utility functions. If the conditions are met, they imply that the dominating utility function will have the higher expected disutility (and aspiration equivalent) for a matching class of probability distributions. Utility dominance relations also provide several insights for the effects of risk aversion and initial wealth on the aspiration equivalent.

**First-Order Utility Dominance**

Let $U_A$ and $U_B$ be two distinct utility functions. We say that $U_A$ utility dominates $U_B$ for all cumulative distributions, $F(x)$, if and only if $U_A(x) \leq U_B(x)$ for all values of $x$, and there is at least some $x_0$ for which strict inequality holds. Now if we are given a lottery having a cumulative distribution, $F(x)$, we have

$$U_A(x) \leq U_B(x) \Rightarrow E_F[DU_A] \geq E_F[DU_B]$$

$\forall$ x with a strict
inequality for at                                        (22)
least one $x_0$.

Where $E_F[DU_A]$ $and$ $E_F[DU_B]$ is the expected disutility for $U_A$ and $U_B$ under the distribution $F(x)$. (The proof follows analogously from stochastic dominance arguments).

Also since the aspiration equivalent is just the inverse of the expected disutility on the cumulative probability distribution function, $F(x)$, we must also have:

$$U_A(x) \leq U_B(x) \Rightarrow \hat{x}_A \geq \hat{x}_B \tag{23}$$

This implies that if a utility function, $U_A$ has first-order utility dominance with respect to another utility function, $U_B$, then for any lottery, decision maker A will always have a higher aspiration equivalent than decision maker, B.

Another implication of equation (22) is

$$U_A(x) \leq U_B(x) \Rightarrow E_F[U_A] \leq E_F[U_B] \tag{24}$$

A utility function, $U_A$, that has first-order utility dominance with respect to another utility function, $U_B$, will thus have a lower expected utility for any given lottery. Utility dominance does not imply that $U_B$ will have a lower certain equivalent. However, in the case of exponential utility functions, we have

$$U_A(x) \leq U_B(x) \Rightarrow \gamma_A \leq \gamma_B \Rightarrow \tilde{x}_A \geq \tilde{x}_B \qquad (25)$$

where $\gamma$ is the risk aversion coefficient, and $\tilde{x}$ is the certain equivalent.

For two exponential utility functions, $U_A$ and $U_B$, we thus have

$$\begin{aligned}
U_A(x) \leq U_B(x) &\Rightarrow E_F[U_A] \leq E_F[U_B] \\
&\Rightarrow \hat{x}_A \geq \hat{x}_B \\
&\Rightarrow \gamma_A \leq \gamma_B \\
&\Rightarrow \tilde{x}_A \geq \tilde{x}_B
\end{aligned} \qquad (26)$$

Equation (26) again shows that higher aspirations are associated with higher risk tolerances (lower risk aversions).

Also if $U_A$ has first-order utility dominance with respect to $U_B$ then the first moment of $U_A$ must be greater than the first moment of $U_B$. (This also follows by analogy where first-order stochastic dominance implies mean dominance for probability functions). The first moment of the utility density function can be determined graphically using the equal areas methods in a manner similar to the graphical determination of the mean for a cumulative distribution function.

On the other hand, if we consider a normalized logarithmic utility function with initial wealth w, over a given domain, we find that higher levels of initial wealth have first-order utility dominance over lower ones. This in turn implies that the aspiration equivalent increases for a given lottery as the decision maker's initial wealth increases.

Using the duality isomorphism, we can also define second-order utility dominance conditions for concave cumulative distributions, S-shaped utility dominance for unimodal probability density functions, and Markowitz (bimodal) utility dominance for cumulative distributions with three inflection points (bimodal probability densities). For a discussion on the latter types of stochastic dominance relations, we refer the reader to Levy and Levy (2002).

## Conclusions

We have presented the concept of duality between probability distributions and utility functions and explored many of its ramifications. The concept of an aspiration equivalent arises naturally as the dual to the certain equivalent, which we used to develop a new method for choosing between lotteries. Because descriptive treatments of decision making often focus on aspiration levels and targets, our work provides a bridge between expected utility and target-based approaches. Specifically, we have shown how aspiration equivalent targets provide a win-win situation for principal-agent delegation and how they should be adjusted in the face of new information. We also developed the relation between the aspiration equivalent and the risk preference of the decision maker. Then we explored the dual problem of finding the decision maker who will set the highest aspiration equivalent for a given lottery. Lastly, we have tied duality to a game theory formulation and illustrated the saddle-point property.

Duality converts many properties and theorems about probability and utility into new ones with their roles reversed. We have examined many dual properties and, in particular, we have introduced new utility dominance theorems as a demonstration of the power of duality. Many of the theorems that have been provided by Pratt (1964) can also be applied in this dual formulation to yield further insights and dual interpretations.

The calculation of an aspiration equivalent leads to deeper questions. Should utility be fixed over time for a person or an organization? Or is risk preference an "emergent" property of our experience in the world? Should it depend on the kinds of opportunities we face, the nature of our competition, or the incentives provided by the organization we work for? In this paper we have addressed incentives that cause the agent to act effectively with the principal's utility function despite his target-based incentives. We envision future work to set normative incentives in other organizational and individual applications.